\documentclass[11pt,draftcls,onecolumn,peerreview]{IEEEtran}
\ifCLASSINFOpdf
\else
\fi
\usepackage{amsmath}

\usepackage{amssymb,xspace,intmacros,subfigure,amsfonts}
\usepackage{epsfig,epsf,epstopdf,graphicx}

\usepackage[ruled]{algorithm2e}

\newcommand{\yb}{{\textbf{y}}}
\newcommand{\Yb}{{\textbf{Y}}}
\newcommand{\Gb}{{\textbf{G}}}
\newcommand{\gb}{{\textbf{g}}}
\newcommand{\sbb}{{\textbf{s}}}
\newcommand{\Sbb}{{\textbf{S}}}

\newcommand{\ab}{{\textbf{a}}}
\newcommand{\bb}{{\textbf{b}}}

\newcommand{\eb}{{\textbf{e}}}
\newcommand{\Ab}{{\textbf{A}}}
\newcommand{\nb}{{\textbf{n}}}
\newcommand{\Nb}{{\textbf{N}}}
\newcommand{\Ib}{{\textbf{I}}}

\begin{document}
\title{Joint DOA Estimation and Array Calibration Using Multiple Parametric Dictionary Learning}

\author{Hamed~Ghanbari,
Hadi~Zayyani,
        and~Ehsan~Yazdian,~\IEEEmembership{Member,~IEEE}% <-this % stops a space
\thanks{H. Ghanbari is with Electrical Engineering Department, Isfahan University of Technology (IUT), Isfahan, Iran, (e-mail:h.ghanbari@ec.iut.ac.ir).}% <-this %
\thanks{H. Zayyani is with the Department
of Electrical and Computer Engineering, Qom University of Technology, Qom, Iran (e-mail: zayyani@qut.ac.ir).}% <-this %
\thanks{E. Yazdian is with Electrical Engineering Department, Isfahan University of Technology (IUT), Isfahan, Iran, (e-mail:yazdian@cc.iut.ac.ir).}

%\author{Author~1,
%Author~2, and~Author~3% <-this % stops a space
%\thanks{Author~1 is with institution 1.}% <-this %
%\thanks{Author~2 is with institution 2.}
%\thanks{Author~3 is with institution 1.}

%stops a space
%\thanks{Manuscript received April 19, 2005; revised December 27, 2012.}
}

\maketitle

\begin{abstract}
	This letter proposes a multiple parametric dictionary learning algorithm for direction of arrival (DOA) estimation in presence of array gain-phase error and mutual coupling. It jointly solves both the DOA estimation and array imperfection problems to yield a robust DOA estimation in presence of array imperfection errors and off-grid. In the proposed method, a multiple parametric dictionary learning-based algorithm with an steepest-descent iteration is used for learning the parametric perturbation matrices and the steering matrix simultaneously. It also exploits the multiple snapshots information to enhance the performance of DOA estimation. Simulation results show the efficiency of the proposed algorithm when both off-grid problem and array imperfection exist.
\end{abstract}

\begin{IEEEkeywords}
Direction of Arrival, Array calibration, Dictionary learning, Steepest-descent.
\end{IEEEkeywords}

 \ifCLASSOPTIONpeerreview
 \begin{center} \bfseries EDICS: SAM-DOAE, MLSAS-SPARSE \end{center}
 \fi
\IEEEpeerreviewmaketitle

\section{Introduction}

\IEEEPARstart{D}{irection} of arrival (DOA) estimation is a famous problem which has various applications in wireless communications \cite{Goda97}, radar \cite{Grec09} and sonar \cite{Thomp93}. There are some classical algorithms for DOA estimation which conventional beamformer \cite{book1}, Minimum Variance Distortionless Response (MVDR) \cite{Capon69} and MUSIC \cite{Schm86} are a few of them. Sparsity-based algorithms are also proposed for DOA estimation which exploit the spatial sparsity of the sources in a discrete grid \cite{Zheng13}-\cite{Gurb12}.

The above mentioned algorithms suffer from the problem of non-calibrations of the array. These array imperfections mainly are gain-phase error, mutual coupling and sensor location errors. In the literature, many algorithms are suggested to jointly estimate the DOA's when these imperfections exist \cite{Ng95}-\cite{Liu16}. Gain-phase error calibration is discussed in \cite{Ng95}-\cite{Liu17}, while mutual coupling calibration is investigated in \cite{Lander91}-\cite{Mao14}. All of these imperfections are regarded in a unified manner in \cite{Ng96}-\cite{Liu16}, comprehensively. In \cite{Ng96}, a Maximum likelihood (ML) estimation algorithm is used for sensor-array calibration. Moreover, \cite{Liu13} suggested a unified framework and a sparse Bayesian method to realize array calibration and DOA estimation simultaneously. In a recent work \cite{Liu16}, a sparse based approach for joint estimation of DOAs and array perturbations is proposed which is based on the sparse assumption of the perturbation matrix. Recently, a blind signal separation method is suggested for joint DOA estimation and array calibration \cite{Liu17}.

In this paper, we treat gain-phase error and mutual coupling in a unified manner. Moreover, similar to \cite{Zamani16}, a dictionary learning algorithm is proposed to solve the DOA estimation problem when there are two aforementioned array imperfection errors. Here, we not only learn the steering dictionary for solving the off-grid problem, but also learn the parametric perturbation matrices to calibrate the array. So, we nominate our proposed algorithm as multiple dictionary learning. In addition, the other novelty in this paper is that in \cite{Zamani16}, only one snap-shot is used for DOA estimation and the estimations from different snapshots are averaged. In this work, we use simultaneously Orthogonal Matching Pursuit (SOMP) \cite{SOMP} in the sparse recovery steps. Hence, we exploit the joint information from all snapshots. Moreover, a benefit of the proposed algorithm is that it uses simple steepest-descent iteration to learn the perturbation models, while the competing state-of-the art algorithms such as sparse-based approach \cite{Liu16} uses more complex convex optimization problems which has no specific simple solution. Besides, in contrast to sparse-based algorithm, we have no further assumption about the sparsity of the perturbation matrices specially about the Mutual Coupling Matrix (MCM). Eventually, simulation results show the superiority of the proposed dictionary learning algorithm over the sparse-based approach, while its computational cost is lower than the aforementioned algorithm.

\section{System model and problem formulation}
\label{sec: Alg}
\subsection{Ideal Array Model}
For the general model of DOA estimation, assume that $K$ far-field sources in direction angles of $\theta_k,k=1,\cdots,K$ in far-field impinging independent narrowband signals $s_k(t)$ into an array in an isometric environment. The array is assumed to be a Uniform Linear Array (ULA) with $M$ omni-directional sensor placed in a line with uniform distribution known as Uniform Linear Array (ULA). The output vector of the array $\mathbf{y}(t)=[y_1(t),\cdots,y_M(t)]^T$ at each time snapshot $t$ can be modeled as:
\begin{equation}
\mathbf{y}(t)=\mathbf{\tilde{A}(\theta)\tilde{s}}(t)+\mathbf{n}(t)
\end{equation}
where $\mathbf{\tilde{s}}(t)=[s_1(t),\cdots,s_K(t)]^T$ is the source vector and  $\mathbf{n}(t)=[n_1(t),\cdots,n_M(t)]^T$ is the sensor array noise vector. The array manifold matrix is $\mathbf{\tilde{A}(\theta)}=[\mathbf{a}(\theta_1),\cdots,\mathbf{a}(\theta_K)]_{M\times K}$ and $\mathbf{a}(\theta_k)=[1,e^{-j\frac{2\pi}{\lambda}d\sin(\theta_k)},\cdots,e^{-j\frac{2\pi}{\lambda}(M-1)d\sin(\theta_k)}]^T$ is the steering vector which provides the delay information of the $k$th source to the all sensors based on the geometry of the array.
The parameter $d$ is the distance between adjacent elements and $\lambda=\frac{c}{f}$ represents the wavelength corresponding to frequency $f$, and $c$ is the velocity of wave propagation. The array manifold $\mathbf{\tilde{A}(\theta)}$ include $K$ columns of steering vectors related to $K$ sources. By discretizing the spatial space into finite angle points and settle the related steering vectors of nonexistent of sources angles into the array manifold, the extended array manifold is obtained and also by extending the vector $\mathbf{s}(t)$ by adding zeros corresponding to the nonexistent source angles, the sparse form of the problem is formulated as
\begin{equation} \label{eq:CS_DOA_prob}
\mathbf{y}(t)=\mathbf{A(\theta)\sbb}(t)+\mathbf{n}(t),
\end{equation}
where $\mathbf{A(\theta)}$ is ${M\times L}$ extended array manifold and $\mathbf{s}(t)$ is $L\times 1$ extended source vector. $L$ is the number of finite angle points in the grids such that $L\gg K$ and $\mathbf{s}(t)$ is $K$-sparse which means only $K$ elements of it is nonzero.\\
\subsection{Array perturbation model}
When the array is not well calibrated, there are imperfections which leverage the DOA estimation performance. In this paper, we focus on the gain-phase error and mutual coupling. In the presence of array perturbations, the received array signal obeys the following model \cite{Liu16}:
\begin{equation} \label{eq:CS_DOA_prob1}
\mathbf{y}(t)=\Gb\mathbf{A(\theta)\sbb}(t)+\mathbf{n}(t),
\end{equation}
where $\Gb$ is the array perturbation matrix which is a parametric dictionary. Following \cite{Liu16}, the matrix $\Gb$ is nominated as $\Gb_{\mathrm{gain}}$, $\Gb_{\mathrm{mutual}}$, in the cases of gain-phase error and mutual coupling, respectively. Collecting all the measurements snapshots in one matrix $\Yb=[\yb(1)|\yb(2)|...|\yb(T)]$, results in the following model:
\begin{equation}
\Yb=\Gb\Ab\Sbb+\Nb
\end{equation}
where $\Sbb=[\sbb(1)|\sbb(2)|...|\sbb(T)]$ is the source matrix and $\Nb=[\nb(1)|\nb(2)|...|\nb(T)]$ is the noise matrix.

The gain-phase error matrix $\Gb_{\mathrm{gain}}$ is a diagonal matrix whose diagonal elements are $g_i=a_i\mathrm{e}^{j\psi_i}$ for $2\le i\le M$ with assuming $g_1=1$ \cite{Liu16}. The Mutual Coupling Matrix (MCM) $\Gb_{\mathrm{mutual}}$ is a toeplitz matrix with the first row equal to $[1, b_1, b_2, ..., b_{M-1}]$, where $b_i$ denotes the complex mutual coupling coefficient between two elements of the array with distance $i$. In \cite{Liu16}, it is assumed that $b_i=0$ for $i\ge3$ which means that only two mutual coefficients are considered. So, then the MCM has a sparse structure. In this paper, we relax this condition and do not assume any constraint on the MCM. However, for the sake of simplicity to derive a closed formulation, we regard only three mutual coupling coefficients and for our simulations only three coefficients are considered. But, there is no systematic restriction to consider more coupling coefficients.

% For the location error of $i$'th sensor we use $\Delta_i$, where it is assumed that $\Delta_0=0$. Due to these location errors, the steering vector can be formulated as
%\begin{equation}
%\ab^{'}(\theta_k)=[1, \mathrm{e}^{\frac{-j2\pi(d+\Delta_1)\sin\theta_k}{\lambda}}, ..., \mathrm{e}^{\frac{-j2\pi((M-1)d+\Delta_{M-1})\sin\theta_k}{\lambda}}]^T.
%\end{equation}
%where the new steering matrix with error locations is $\Ab^{'}(\theta)=[\ab^{'}(\theta_1), ..., \ab^{'}(\theta_L)]$. For simplicity, in this paper, for the location error perturbation, we use the following model:
%\begin{equation} \label{eq:CS_DOA_prob2}
%\mathbf{y}(t)=\Ab^{'}(\theta, \Delta)\sbb(t)+\mathbf{n}(t),
%\end{equation}
%where $\Delta=[\Delta_1,..., \Delta_{M-1}]^T$ is the error location vector.

\section{The proposed algorithm-one snapshot case}\label{sec: Alg}
In this section, we introduce a multiple parametric dictionary learning technique for grid mismatch problem of estimated DOAs in presence of two aformentioned types of imperfections. The basic idea, is to learn the parametric dictionaries in the model (\ref{eq:CS_DOA_prob1}). In the learning steps of the proposed algorithm, we use the cost function of $J=||\mathbf{y}-\Gb\mathbf{A(\theta)\sbb}||^2_2$. The overall algorithm is a three step iterative algorithm.

The first step is to recover the sparse vector $\sbb$ assuming that the dictionaries $\Gb$, $\Ab(\theta)$ are fixed. This step is done by OMP algorithm in the one snapshot case and with a SOMP algorithm \cite{SOMP} in the case of multiple snapshots.

The second step is to learn the parameters of the dictionary $\Ab(\theta)$ which are $\theta_k$s, assuming that the perturbation matrix $\Gb$ is fixed and it is done for a number of iterations. In the cases of gain-phase error and mutual coupling, the second step which presents a solution for off-grid problem, is similar to those suggested in \cite{Zamani16}. Since, here, we have a perturbation matrix $\Gb$ in the model (\ref{eq:CS_DOA_prob1}), there is a small difference in updating the angles in comparison to \cite{Zamani16}. Now, we use an steepest-descent algorithm for updating $\theta_k$ for minimizing the cost function $J=||\mathbf{y}-\Gb\mathbf{A(\theta)\sbb}||^2_2$, assuming $\Gb$ and $\sbb$ are fixed. Similar calculations to those presented in \cite{Zamani16}, show that the final recursion for updating the angles are:
\begin{equation}
\theta_k^{\mathrm{new}}=\theta_k^{\mathrm{old}}+\mu_{\theta}\mathrm{Re}\{c_0s_k\eb^{H}\Gb(\ab(\theta_k)\odot\bb(\theta_k))\}
\end{equation}
where $\mu_{\theta}$ is the step-size, $c_0=j\frac{2\pi d}{\lambda}$, $\eb=\Gb\mathbf{A(\theta)\sbb}-\yb$ is the error vector, $\ab(\theta_k)$ is the steering vector, and $\bb(\theta_k)=[0\quad\cos(\theta_k) ...\quad (M-1)\cos(\theta_k)]^T$. Since we have only one snapshot, we omitted the index $t$ in the equations in this section.\\

%In the case of sensor location error, updating the angles based on the steepest-descent to minimize the cost function $J=||\mathbf{y}(t)-\Ab^{'}(\theta, \Delta)\sbb||^2_2$, assuming $\mathbf{\theta}$ and $\mathbf{\Delta}$ are fixed, results to the following formula:
%\begin{equation}
%\theta_k^{\mathrm{new}}=\theta_k^{\mathrm{old}}+\mu_{\theta}\mathrm{Re}\{c_0s_k\eb^{H}(\ab^{'}(\theta_k)\odot\bb^{'}(\theta_k))\}
%\end{equation}
%where $\bb^{'}(\theta_k)=[0\quad(d+\Delta_1)\cos(\theta_k) ...\quad ((M-1)d+\Delta_{M-1})]^T$.\\

The third step, is to learn the parameters of the perturbation matrix assuming that the sparse vector and the angles $\theta$ are fixed and it is done for a number of iterations. For the third step, we drive the proposed learning formula of parameters using steepest-descent in the following two cases.
\subsection{Gain-phase error perturbation}
When gain-phase error exists, at the third step of the algorithm, we want to update or learn the parameters $g_i$. We use a simple steepest-descent algorithm. The iteration to update the $g_i$ is $g^{new}_i=g^{old}_i-\mu_g\frac{\partial J}{\partial g_i}$, where $J=||\mathbf{y}-\Gb_{\mathrm{gain}}\mathbf{\phi}||^2_2$ and $\phi=\mathbf{A(\mathbf{\theta})\sbb}$ is assumed known and fixed in this step of the algorithm. If we define the error as $\eb=\yb-\Gb\mathbf{\phi}$, then the cost function is defined as $J=\eb^H\eb$. Employing partial derivatives $\frac{\partial (\eb^H\eb)}{\partial g_i}=\frac{\partial (\yb^H\yb)}{\partial g_i}-\frac{\partial (\yb^H\Gb\mathbf{\phi})}{\partial g_i}-\frac{\partial ((\Gb\phi)^H\yb)}{\partial g_i}+\frac{\partial ((\Gb\phi)^H\Gb\mathbf{\phi})}{\partial g_i}$, and simple manipulations, we have the final recursion formula for updating $\gb=[g_1,...,g_M]^T$:
\begin{equation}
\label{eq: gp}
\gb^{new}=\gb^{old}+\mu_g(\yb^{*}\phi-(\gb^{old})^{*}\phi^{*}\phi),
\end{equation}
where $\mu_g$ is the step-size of the steepest-descent algorithm.

\subsection{Mutual coupling}
\label{sec: Mutual}
In this subsection, we should obtain the update formula for the coupling coefficients. Similarly, we use steepest-descent to update these coefficients. Hence, we have $b^{new}_i=b^{old}_i-\mu_b\frac{\partial J}{\partial b_i}$ for $1\le i\le M-1$. For simplicity of deriving the closed formula, we assume that just three coefficients are non-zero. The generalization of the formulas for other cases is straightforward. It is an advantage of the proposed algorithm over sparse-based approach \cite{Liu16} where it only uses two nonzero coefficients. Based on defining the cost function as $J=||\mathbf{y}-\Gb_{\mathrm{mutual}}\phi||^2_2$, we have $J= (\mathbf{y}-\Gb_{\mathrm{mutual}}\phi)^H(\mathbf{y}-\Gb_{\mathrm{mutual}}\phi) =\mathbf{y}^H\mathbf{y}-\mathbf{y}^H\Gb_{\mathrm{mutual}}\phi-(\Gb_{\mathrm{mutual}}\phi)^Hy+(\Gb_{\mathrm{mutual}}\phi)^H(\Gb_{\mathrm{mutual}}\phi)$. The gradient of $J$ is calculated as follows $\nabla_{b} (J)=-\nabla_{b}(\mathbf{y}^H\Gb_{\mathrm{mutual}}\phi)-\nabla_{b}((\Gb_{\mathrm{mutual}}\phi)^H\mathbf{y})+\nabla_{b}((\Gb_{\mathrm{mutual}}\phi)^H(\Gb_{\mathrm{mutual}}\phi))
=\mathbf{\Psi}\mathbf{y}^*-\mathbf{\Psi}\mathbf{w}^*$. Simplifying is done by using the following definitions $\mathbf{\omega}=\Gb_{\mathrm{mutual}}\mathbf{\phi}$\\
1. $-\frac{\partial (\mathbf{y}^H\Gb_{\mathrm{mutual}}\phi)}{\partial b_{i}}=-\frac{\partial \omega}{\partial b_{i}}\mathbf{y}^*$, $\frac{\partial \omega}{\partial b_{i}}=[ \frac{\partial \omega_{1}}{\partial b_{i}}, ..., \frac{\partial \omega_{M}}{\partial b_{i}}]$\\
2. $-\frac{\partial ((\Gb_{\mathrm{mutual}}\phi)^Hy)}{\partial b_{i}}=0$\\
3. $\frac{\partial((\Gb_{\mathrm{mutual}}\phi)^H(\Gb_{\mathrm{mutual}}\phi))}{\partial b_{i}} =\frac{\partial (\omega\omega^H)}{\partial b_{i}}=\sum_{j=1}^M\frac{\partial (\omega\omega^H)}{\partial \omega_{j}}\frac{\partial \omega_{j}}{\partial b_{i}}=\sum_{j=1}^M\omega_{j}^*\frac{\partial (\omega_{j})}{\partial b_{i}}$.\\
%$\mathbf{y}^H\Gb_{\mathrm{mutual}}\phi=y_{1}w_{1}+y_{2}w_{2}+  \cdots + y_{M}w_{M}$\\
So, the final recursion for updating the mutual coefficient vector $\bb=[b_1,b_2,b_3]^T$ is
\begin{equation}
\label{eq: mutual}
\bb^{new}=\bb^{old}+\mu_b(\Psi(\omega-\yb)^{*}),
\end{equation}
where $\mathbf{\omega}=\Gb_{\mathrm{mutual}}^{old}\Ab\sbb$, and $\mathbf{\Psi}$ is equal to
\begin{equation}
\label{eq: psi}
\mathbf{\Psi}=\begin{pmatrix} \phi_2 & \phi_1+\phi_3 & \phi_2+\phi_4 & \phi_3+\phi_5 & \ldots & \phi_{M-1}\\ \phi_3 & \phi_4 & \phi_1+\phi_5 & \phi_2+\phi_6 & \ldots & \phi_{M-2}\\ \phi_4 & \phi_5 & \phi_6 & \phi_3+\phi_7 & \ldots & \phi_{M-3} \end{pmatrix}.
\end{equation}

\section{The proposed algorithm-multiple snapshot case}\label{sec: Alg1}
The drawback of the dictionary learning-based algorithm presented in \cite{Zamani16} is that it uses only one snapshot and average the results of estimations in the multiple snapshots. In the case of multiple snapshots, for the sparse recovery, we use the SOMP algorithm \cite{SOMP} to recover $\Sbb$ based on \Yb, $\Gb$ and $\Ab(\theta)$. Inspiring from \cite{Zayy16}, here, for the case of multiple snapshots, we employ $J=\sum_{t=1}^T||\mathbf{y}(t)-\Gb\Ab\sbb(t)||^2_2$ as the cost function for both gain-phase error case and mutual coupling case, where $T$ is the number of snapshots. Since the multiple snapshot cost function is the sum of the one snapshot case, the error update of the recursion of each parameter is the addition over all snapshots. Therefore, we have the following formulas for updating the parameters of the perturbation matrices and updating the off-grid angles:
\begin{equation}
\label{eq: gpmul}
%g^{new}_i=g^{old}_i+\mu_g\sum_{t=1}^T[y_i(t)^{*}\phi_i(t)-(g^{old}_i)^{*}\phi_i(t)^{*}\phi_i(t)],
\gb^{new}=\gb^{old}+\mu_g\sum_{t=1}^T(\yb^{*}(t)\mathbf{\phi}(t)-(\gb^{old})^{*}\mathbf{\phi}^{*}(t)\mathbf{\phi}(t)),
\end{equation}
\begin{equation}
\label{eq: mutualmul}
\bb^{new}=\bb^{old}+\mu_b\sum_{t=1}^T\mathbf{\Psi}(t)(\mathbf{\omega}(t)-\yb(t))^{*},
\end{equation}
%\begin{equation}
%	\label{eq: loc}
%	 \mathbf{\Delta}^{new}=\mathbf{\Delta}^{old}+\mu_{\Delta}\sum_{t=1}^T\mathrm{Re}\{\mathrm{diag}(\eb^{H}(t))\Ab^{'}(\mathbf{\theta},\mathbf{\Delta})(s_k(t)\odot\zb)\}
%\end{equation}
\begin{equation}
\theta_k^{\mathrm{new}}=\theta_k^{\mathrm{old}}+\mu_{\theta}\sum_{t=1}^T\mathrm{Re}\{c_0s_k(t)\eb^{H}(t)\Gb(\ab(\theta_k)\odot\bb(\theta_k))\}
\end{equation}
%\begin{equation}
%\theta_k^{\mathrm{new}}=\theta_k^{\mathrm{old}}+\mu_{\theta}\sum_{t=1}^T\mathrm{Re}\{c_0s_k(t)\eb^{H}(t)(\ab^{'}(\theta_k)\odot\bb^{'}(\theta_k))\}
%\end{equation}
where $\mathbf{\phi}(t)=\mathbf{A(\mathbf{\theta})\sbb(t)}$, $\mathbf{\Psi}(t)$ is defined in (\ref{eq: psi}), $\mathbf{\omega}(t)=\Gb_{\mathrm{mutual}}^{old}\Ab(\mathbf{\theta})\sbb(t)$ and $\eb(t)=\yb(t)-\Gb\mathbf{\phi}(t)$. The details of the three-step multiple dictionary learning algorithm in multiple snapshots case are illustrated in Algorithm \ref{MDL}. The One snapshot case is the same as multiple snapshots case where $T=1$.

\begin{algorithm}
	\label{MDL}
	\SetKwData{Left}{left}
	\SetKwData{This}{this}
	\SetKwData{Up}{up}
	\SetKwFunction{Union}{Union}
	\SetKwFunction{FindCompress}{FindCompress}
	\SetKwInOut{Input}{input}
	\SetKwInOut{Output}{output}
	\caption{Multiple dictionary learning Algorithm}
	\Input{\\
		Array output snapshots\\ $\mathbf{Y}=\Gb\Ab(\theta)\Sbb+\Nb\in \mathbb{R}^{M\times T}$\\
		%Array manifold matrix $\mathbf{A(\theta)} \in \mathbb{C}^{M\times L}$\\
		%Initial values for $\mathbf{A}$ and $\mathbf{S}$\\
	}
	\Output{\\
		DOA estimation based on $\widehat{\Sbb}\in \mathbb{R}^{L\times T}$, $\hat{\Gb}$, and $\hat{\theta}$}
	$\mathrm{Initialization}:\Gb^0=\Ib,\theta^{0}=\mathrm{Uniform}[-90,90]$;\\
	%\For{$l=1$ \KwTo $L$}{
	%Initialize $\mathbf{A}_l$ and $\mathbf{S}_l$ using VCA for the first layer and using random initialization for other layers\\
	\For{$k_{1}=1$ \KwTo $\mathrm{Iter1}_{max}$}{
		%update $\mathbf{A}_l$ using (\ref{updatea})\;
		$\mathrm{STEP 1}:\mathrm{Learning}\quad\Sbb$\\
		$\Sbb^k=\mathrm{SOMP}(\Yb,\Ab,\Gb))$\;
		$\mathrm{STEP 2}: \mathrm{Learning}\quad\theta$\\
		\For{$k_{2}=1$ \KwTo $\mathrm{Iter2}_{max}$}{
			$\mathrm{Update}\quad\theta\quad\mathrm{Based}\quad\mathrm{on}\quad(11)$\;
		}
		$\mathrm{STEP 3}:\mathrm{Learning}\quad\Gb$\\
		\For{$k_{3}=1$ \KwTo $\mathrm{Iter3}_{max}$}{
			$\mathrm{Update}\quad\Gb\quad\mathrm{Based}\quad\mathrm{on}\quad(9),(10)$\;
		}
		%$\mathbf{X}_l=\widetilde{\mathbf{X}}_l$ , $\mathbf{A}_l=\widetilde{\mathbf{A}}_l$ using (\ref{fcls})\;
		%update $\mathbf{S}_l$ using (\ref{updates})\;
		%\If{stopping criteria in (\ref{stop})}{break}
	}
	%$\mathbf{X}_{l+1}=\mathbf{S}_l$
	
	$\widehat{\Sbb}\gets \Sbb^{\mathrm{Iter}_{max}}$, $\widehat{\Gb}\gets \Gb^{\mathrm{Iter}_{max}}$, $\widehat{\theta}\gets \theta^{\mathrm{Iter}_{max}}$;
	%}
	%$\mathbf{A}=\prod_{l=1}^{L}\mathbf{A}_l$ and $\mathbf{S}=\mathbf{S}_{L}$
	
\end{algorithm}
%\begin{equation}
%\mathrm{Par}^{new}=\mathrm{Par}^{old}-\mu_{\mathrm{Par}}\sum_{t=1}^T\frac{\partial ||\mathbf{y}(t)-\Gb\Ab\sbb(t)||^2_2}{\partial \mathrm{Par}}.
%\end{equation}
%Therefore, in multiple snapshot case, the correction terms in the recursions (\ref{eq: gp}) and (\ref{eq: mutual}) are summed over multiple snapshots. For brevity, we omit the new recursions in the multiple snapshot case.
\section{Simulation Results}
\label{sec: Sim}
This section presents the simulation results. In the simulations, two experiments were performed to show the efficiency of the proposed multiple dictionary learning based DOA estimation algorithms. We considered three sources ($K=3$) at angles $\theta_1=-12.50^\circ$, $\theta_2=43.85^\circ$ and $\theta_3=76.80^\circ$ and the number of array elements are assumed to be $M=25$. For the discrete grid, the angle interval $[-90^\circ,90^\circ]$ is divided into $91$ equal bins with the step of $2^\circ$. The sensor array signal $\mathbf{s}(t)$ and noise $\mathbf{n}(t)$ are regarded to be complex independent white Gaussian with zero mean. The Signal to Noise Ratio (SNR) is defined as $\mathrm{SNR(dB)}=10\mathrm{log}(\frac{E\{|\Ab\sbb|^2\}}{E\{|\nb|^2\}})$. For the performance metric, Mean Square Error (MSE) of estimated angles is used which is defined as $\mathrm{MSE}=\sqrt{\frac{1}{K}\sum_{i=1}^K(\theta_i-\hat{\theta_i})^2}$. The values of MSE are averaged over $50$ independent monte carlo runs. In the proposed method, the iteration numbers are selected as $\mathrm{Iter1}_{max}=20$, $\mathrm{Iter2}_{max}=40$, and $\mathrm{Iter3}_{max}=40$. For simulating the sparse-based method \cite{Liu16}, number of iterations for outer loop and inner loop are selected as $10$, the other parameters are chosen as $\tau=\frac{1}{25}$ and $\rho=\frac{1}{25}$. All algorithms are compared with the same initial values. The interelement spacing of the ULA is assumed to be $\frac{d}{\lambda}=0.5$.

\begin{figure}[tb]
\begin{center}
\includegraphics[width=8cm]{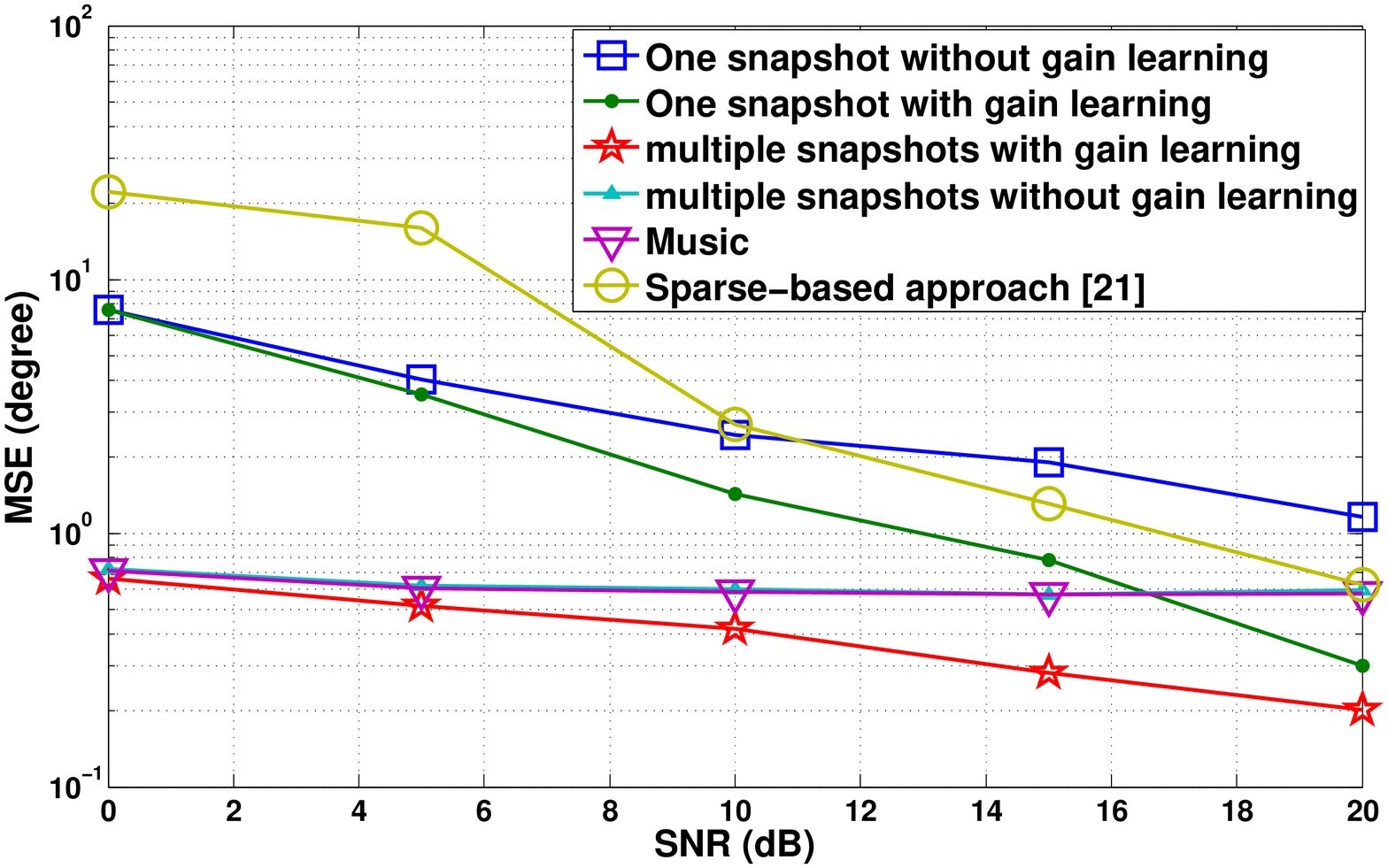}
\end{center}
\caption{MSE versus SNR for DOA estimation in the case of gain-phase error.}
%\end{center}
\label{fig1}
\end{figure}

At the first experiment, similar to \cite{Lander91}, the amplitude and phase is assumed to be $a_i=[(\beta_i-0.5)\sigma_a.\sqrt{12}+1]$ and $\psi_i=[(\gamma_i-0.5)\sigma_{\psi}\sqrt{12}+1]$, where $\beta_i$ and $\gamma_i$ are uniformly distributed between zero and one, and $\sigma_a=0.1$ and $\sigma_{\psi}=2^\circ$. The number of snapshots is considered as $T=5$. In this experiment, we used six algorithms, which are OMP algorithm without dictionary learning (results are averaged for five snapshots), OMP with dictionary learning (results are averaged over five snapshots), SOMP without dictionary learning, SOMP with dictionary learning, sparse-based algorithm \cite{Liu16}, and MUSIC \cite{Schm86}. Therefore, the performance metric of the algorithms (MSE), is calculated at different SNRs. For updating  gain, the step size  in one snapshot and multiple snapshots  are selected as  $\mu_g=1\times10^{-4}$. The step size for learning angle $\mu_{\theta}$ in one snapshot and multiple snapshots cases are chosen as $2\times10^{-5}$ and $1\times10^{-5}$, respectively.    The results are shown in Fig~\ref{fig1}. It shows that the proposed multiple dictionary learning based algorithm is better than the other algorithms.\\ % is better than other algorithms. It also shows that the IMAT is better than OMP, MUSIC and \lone-SVD.\\

To compare the computational cost of the algorithms, in experiment 1, we calculated the average simulation time of the proposed multiple dictionary learning algorithm and sparse-based algorithm for one snapshot case. The times are 3.40 and 27.09 seconds for the proposed algorithm and sparse-based algorithm, respectively. Hence, our proposed algorithm is less complex than sparse-based algorithm while simultaneously has better performance in estimating the DOAs. On the other hand, the average simulation time of MUSIC for 5 snapshots is equal to 0.55 seconds. Therefore, our algorithm is more complex than MUSIC while has better performance than it.

\begin{figure}[tb]
\begin{center}
\includegraphics[width=8cm]{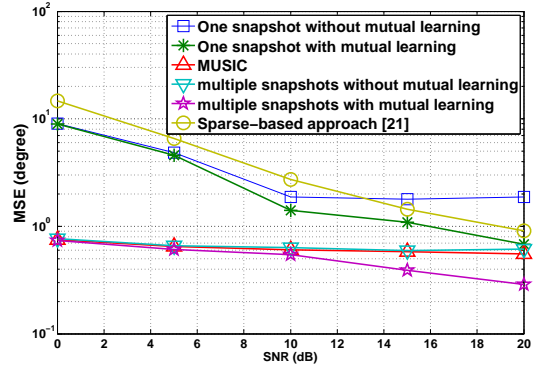}
\end{center}
\caption{MSE versus SNR for DOA estimation in the case of mutual coupling.}
%\end{center}
\label{fig2}
\end{figure}

At the second experiment, the mutual coupling is regarded as the imperfection of the array. Unlike \cite{Liu16} which used two nonzero mutual coupling coefficients, we used three nonzero coefficients. Also, we assumed that the couplings are four times stronger than those used in \cite{Liu16}. So, we selected $b_1=4\times(0.03+j0.077)$, $b_2=4\times(0.016+j0.019)$ and $b_3=4\times(0.036+j0.012)$. For updating mutual, the step size  in one snapshot and multiple snapshots are selected as  $\mu_b=1\times10^{-4}$. The step size for learning angle $\mu_{\theta}$ in one snapshot and multiple snapshots cases are selected as $2\times10^{-5}$ and $5\times10^{-4}$, respectively.  The results are shown in Fig~\ref{fig2}. It shows that the proposed multiple dictionary learning based algorithm outperforms the other algorithms. % is better than other algorithms. It also shows that the IMAT is better than OMP, MUSIC and \lone-SVD.

%\begin{figure}[tb]
%\begin{center}
%%\includegraphics[width=8cm]{res.eps}
%\includegraphics[width=8cm]{fig1_revise.eps}
%\end{center}
%\caption{MSE versus SNR for DOA estimation from the array sensor output.}
%%\end{center}
%\label{fig1}
%\end{figure}
%
%\begin{figure}[tb]
%\begin{center}
%\includegraphics[width=8cm]{onebit_compare_all_methods_new_Labels.eps}
%\end{center}
%\caption{MSE versus SNR for DOA estimation from the sign of the array sensor output.}
%%\end{center}
%\label{fig2}
%\end{figure}

%\begin{figure}[tb]
%\begin{center}
%\includegraphics[width=8cm]{onebit_compare_all_methods_new_Labels.eps}
%\end{center}
%\caption{MSE versus SNR for DOA estimation from the sign of the array sensor output.}
%%\end{center}
%\label{fig3}
%\end{figure}

%\begin{figure}[tb]
%\begin{center}
%\includegraphics[width=8cm]{exp1_final.jpg}
%\end{center}
%\caption{NMSE versus number of training signals.}
%%\end{center}
%\label{fig2}
%\end{figure}
%
%\begin{figure}[tb]
%\begin{center}
%\includegraphics[width=8cm]{exp2_final.jpg}
%\end{center}
%\caption{NMSE versus number of sign measurements.}
%%\end{center}
%\label{fig3}
%\end{figure}

\section{Conclusion}
\label{sec: con}
We have proposed new iterative multiple parametric dictionary learning based algorithm for DOA estimation in presence of off-grid error and two types of array imperfection. The imperfections are gain-phase errors and mutual coupling. In these cases, the steepest-descent is used to update the parametric perturbation dictionaries as well as steering matrix. Simulation results show that multiple parametric dictionary learning algorithm outperforms some other algorithms such as MUSIC and sparse-based algorithm \cite{Liu16}, while its complexity is less than sparse-based approach and higher than MUSIC.

%A Bayesian hypothesis test is proposed to detect the active elements of a sparse vector in one bit compressed sensing framework. Then, the amplitudes of active elements is obtained by an ML estimator. Simulation results in a special case, show that using new algorithm improves the accuracy of sparse vector estimation by at least 4dB.


%\appendices
%\section{Final recursion of gain-phase coefficients}
%\label{sec: app1}
%a
%\section{Final recursion of mutual coupling coefficients}
%\label{sec: app2}
%For one-bit DOA estim
%
%\section{Final recursion of sensor location errors}
%\label{sec: app2}
%For one-bit DOA estim

%\section*{Acknowledgment}

\ifCLASSOPTIONcaptionsoff
  \newpage
\fi

\end{document}